\documentclass[letterpaper,journal]{IEEEtran}
\usepackage{amsmath,amsfonts}
\usepackage{algorithmic}
\usepackage{algorithm}
\usepackage{array}
\usepackage[caption=false,font=scriptsize,labelfont=sf,textfont=sf]{subfig}
\usepackage{textcomp}
\usepackage{stfloats}
\usepackage{url}
\usepackage{verbatim}
\usepackage{graphicx}
\usepackage{cite}
\usepackage{multirow}
\usepackage{tabularx}
\usepackage{color}
\usepackage[colorlinks,bookmarksopen,bookmarksnumbered,citecolor=green, linkcolor=red, urlcolor=blue]{hyperref}
\usepackage{bm}
\usepackage{makecell}
\usepackage{bigstrut}
\usepackage{siunitx}
\usepackage{enumerate}
\usepackage{booktabs}
\usepackage{orcidlink}

\providecommand{\best}[1]{\textbf{#1}}
\definecolor{secondblue}{rgb}{0.05,0.30,0.75}
\providecommand{\second}[1]{\textcolor{secondblue}{\textbf{#1}}}

\begin{document}

\title{Flow-based Gaussian Splatting for Continuous-Scale Remote Sensing Image Super-Resolution}

\author{Jiangwei Mo, Xi Lu, Hanlin Wu\,\orcidlink{0000-0002-3505-0521},~\IEEEmembership{Member,~IEEE}    
    \thanks{This work was supported in part by the National Natural Science Foundation of China under Grant 62401064, and in part by the Fundamental Research Funds for the Central Universities under Grant 2026YB042 and Grant 2024TD001. \emph{(Jiangwei Mo and Xi Lu contributed equally to this work. Corresponding author: Hanlin Wu.)}
    }
    \thanks{The authors are with the School of Information Science and Technology, Beijing Foreign Studies University, Beijing 100089, China (e-mail: hlwu@bfsu.edu.cn).}
}

\markboth{}%
{}

\IEEEpubid{}

\maketitle
\begin{abstract}
    High-resolution remote sensing images (RSIs) are crucial for Earth observation applications, yet acquiring them is often limited by sensor constraints and costs. In recent years, generative super-resolution (SR) methods, particularly diffusion models, have made significant progress. However, they typically require slow iterative inference with 40--1000 steps and exhibit limited flexibility in continuous-scale SR settings. To address these issues, we propose FlowGS, a generative reconstruction framework for arbitrary-scale SR of RSIs. FlowGS models the high-frequency detail representations between high- and low-resolution images and learns a continuous probability flow from noise to detail priors via flow matching (FM) constrained by shortcut consistency, thereby reducing generative complexity and improving inference efficiency. Additionally, we employ 2D Gaussian splatting to construct a continuous feature field, thereby enabling flexible reconstruction at arbitrary query locations. Experimental results show that FlowGS delivers competitive perceptual quality compared with existing methods in both continuous-scale and fixed-scale SR settings, with substantially improved inference efficiency.
\end{abstract}


\section{Introduction}

\IEEEPARstart{R}{emote} sensing image super-resolution (RSISR) aims to recover high-resolution (HR) details from low-resolution (LR) observations, thereby enhancing the quality and utility of remote sensing (RS) imagery for Earth observation \cite{rs_sr_review}. Compared with natural images, RS imagery typically exhibits larger spatial coverage, wider scale variations, and more complex structures, posing greater challenges for detail reconstruction and scale generalization.

Continuous-scale SR aims to support arbitrary scale factors within a single model, eliminating the need for scale-specific models. Meta-SR \cite{metasr} first demonstrated its feasibility, and subsequent implicit neural representation (INR)-based methods \cite{liif,lte,funsr,srfeinr} further established continuous coordinate-based decoding as the dominant paradigm. However, these methods still store features as discrete latent codes and rely on implicit interpolation to produce continuous outputs, which may fail to preserve spatial continuity when query locations change and can even introduce artifacts around feature boundaries.

\begin{figure}[t]
    \centering
    \includegraphics[width=\columnwidth]{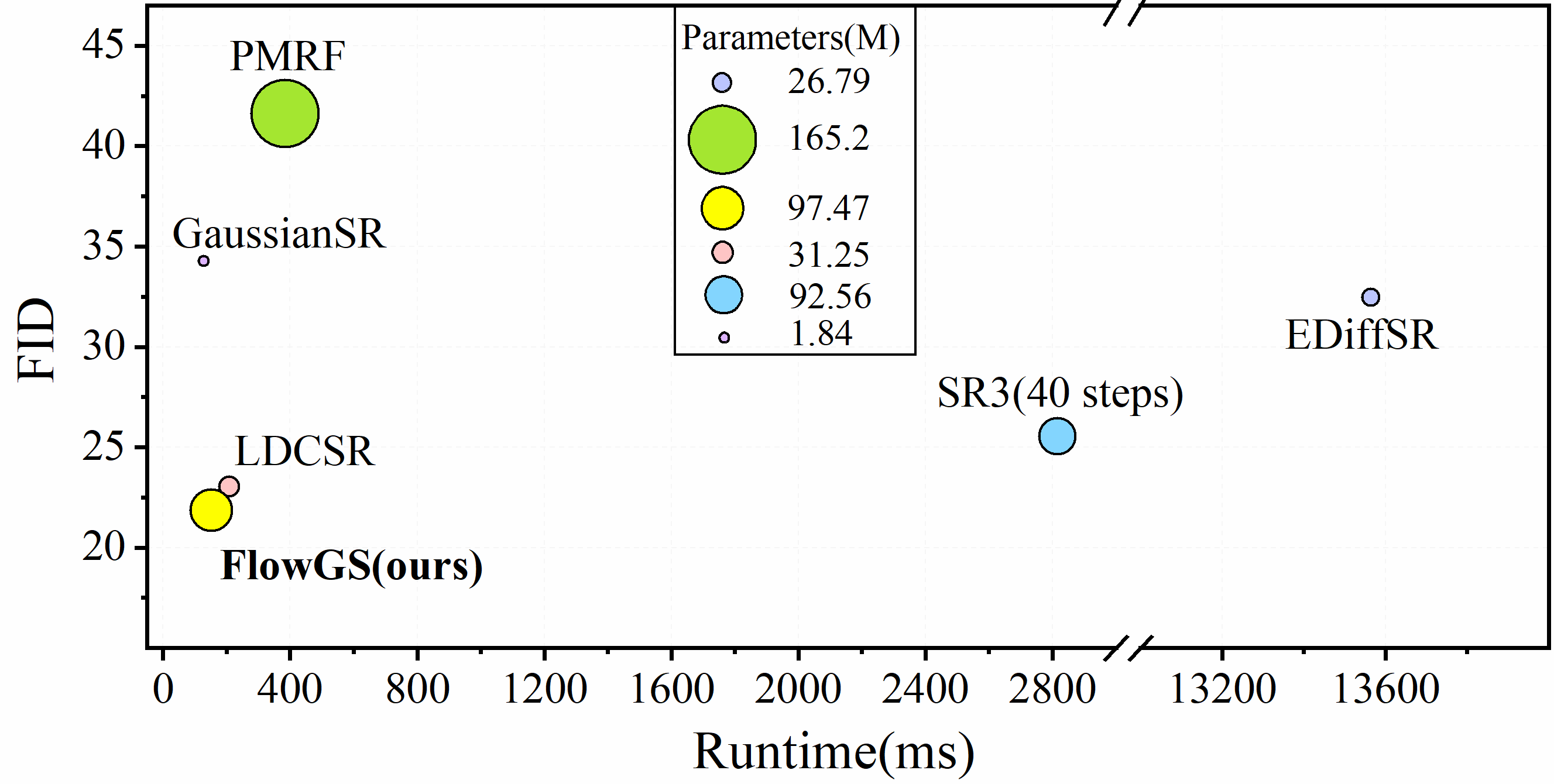}
    \caption{FID is reported for $\times$4 SR on the DIOR dataset, while inference time is measured on $512 \times 512$ images using a single NVIDIA RTX 4090 GPU.}
    \label{fig:efficiency-FID}
\end{figure}

To strengthen spatial continuity, GaussianSR \cite{gaussiansr} introduces 2D Gaussian splatting into continuous-scale SR. By replacing discrete point-wise features with continuous Gaussian fields, GaussianSR allows each query location to aggregate information from multiple overlapping kernels, leading to more coherent feature interpolation and stronger modeling of complex structures and long-range dependencies. However, this improvement lies primarily in representation and rendering: GaussianSR remains essentially a deterministic regression method, lacking an explicit mechanism to model the distribution of missing high-frequency details. As a result, although it improves spatial continuity over INR-based methods, its ability to recover realistic fine textures is still fundamentally constrained.

Generative SR methods offer a complementary advantage by modeling restoration as conditional distribution learning. Diffusion-based SR methods achieve strong perceptual quality \cite{ddpm,sr3,ediffsr,ldm}, but their multi-step denoising process incurs substantial inference cost. Flow matching (FM) \cite{flowmatching} provides a promising alternative to diffusion-style generation by directly learning the velocity field of a transport process, allowing the target distribution to be reached through a continuous path with only a few integration steps.

Motivated by these observations, we propose FlowGS, a generative Gaussian splatting framework driven by FM for continuous-scale RSISR. Specifically, we first represent missing high-frequency details with a compact latent prior and decode it into Gaussian-aware multi-scale features, so that the generated signal is directly aligned with continuous rendering. We then learn a conditional flow transport from noise to this detail prior, replacing diffusion-style iterative denoising with an efficient one-step generation process. Finally, the generated detail prior is injected into a scale-aware reconstruction pipeline and rendered through Gaussian aggregation to produce high-quality super-resolved results across arbitrary scale factors. By unifying continuous-scale modeling, Gaussian-based rendering, and one-step flow-based generation, FlowGS achieves a better trade-off between generation quality and computational efficiency, as illustrated in Fig.\,\ref{fig:efficiency-FID}.

The contributions of this work are threefold.

\begin{enumerate}[1)]
    \item We introduce a Gaussian-oriented detail prior that bridges generative latent modeling and continuous spatial rendering in continuous-scale SR.
    \item We adopt conditional FM in the detail-prior space to replace iterative latent diffusion for more efficient RSISR.
    \item We employ a shortcut-enhanced velocity supervision to promote locally linear transport, enabling stable one-step inference without sacrificing detail fidelity.
\end{enumerate}

\begin{figure*}[t]
    \includegraphics[width=\linewidth]{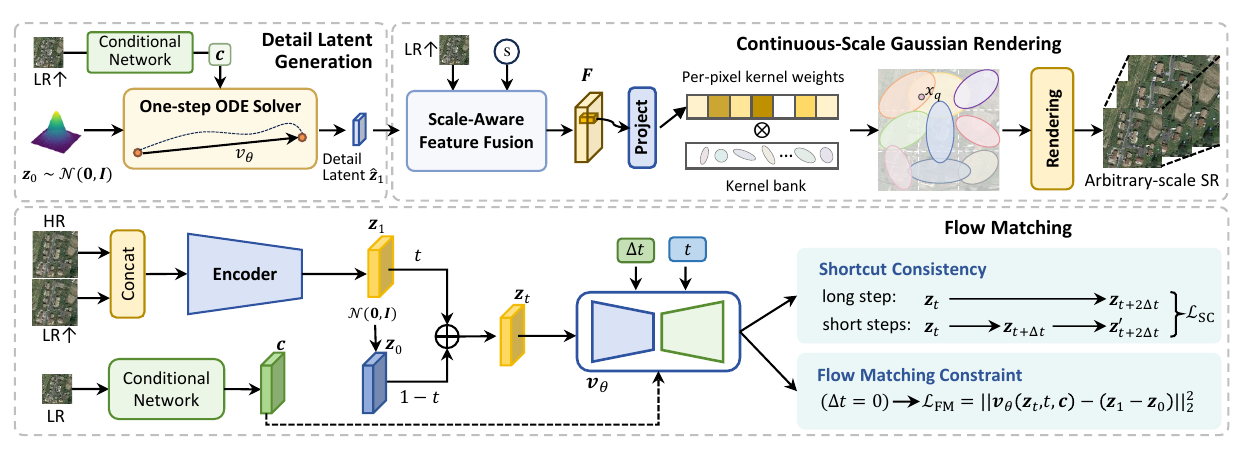}
    \caption{Overview of FlowGS. The upper panel illustrates the inference pipeline, and the lower panel shows the training process of FM.}
    \label{fig:flowgs}
\end{figure*}

\section{Methodology}

Given an LR image $I_{\mathrm{LR}}$ and an arbitrary scale factor $s$, our goal is to reconstruct a super-resolved image $I_{\mathrm{SR}}$ at the target resolution using a single continuous-scale model. As illustrated in Fig.\,\ref{fig:flowgs}, FlowGS consists of two integrated modules: 1) FM-based detail latent generation and 2) continuous-scale Gaussian rendering.

\subsection{FM-based Detail Latent Generation}

SR can be viewed as recovering missing high-frequency details conditioned on the LR input. Instead of generating the full image, we model a compact \emph{detail latent} that captures these missing components. To learn its distribution, we adopt FM in the latent space, which models a continuous transport trajectory from Gaussian noise to the target detail latent via a velocity field.

\subsubsection{Detail Latent Construction}
Given $I_{\mathrm{HR}}$ and $I_{\mathrm{LR}}$, we first upsample $I_{\mathrm{LR}}$ to $I_{\mathrm{LR}}^\uparrow$, where $I_{\mathrm{HR}}, I_{\mathrm{LR}}^\uparrow \in \mathbb{R}^{H \times W \times C_{\mathrm{in}}}$ and $C_{\mathrm{in}}$ is the number of input spectral bands. We define the detail latent by encoding the concatenation of the HR image and the upsampled LR observation:
\begin{equation}
    \label{eq:encoder}
    \bm{z}_1 = \mathcal{E}(\mathrm{Concat}(I_{\mathrm{HR}}, I_{\mathrm{LR}}^\uparrow)) \in \mathbb{R}^{H/8 \times W/8 \times C}.
\end{equation}
Here, $\mathcal{E}$ follows the encoder architecture of LDM~\cite{ldm}, with doubled input channels for the concatenated input. It is pretrained together with the continuous-scale Gaussian rendering module introduced in Sec.\,\ref{sec:continuous-scale-rendering}, so that the latent representation is aligned with the subsequent rendering process.

Since $I_{\mathrm{HR}}$ is unavailable during inference, $\bm z_1$ cannot be directly obtained from the input. We therefore learn to estimate its conditional distribution from the LR observation. To provide structural guidance for this estimation, we extract a conditional embedding from the upsampled LR image:
\begin{equation}
    \bm c=\mathrm{CondNet}(I_{\mathrm{LR}}^\uparrow),
\end{equation}
where $\mathrm{CondNet}$ adopts the same encoder architecture as $\mathcal{E}$ but uses independent weights. The resulting condition $\bm c$ encodes LR structural information and is used to guide the flow-based generation of the target detail latent.

\subsubsection{Conditional Flow Transport}

With the target detail latent $\bm z_1$ and the LR structural condition $\bm c$ defined above, we adopt FM to model the conditional distribution of $\bm z_1$. FM learns a continuous transport from a simple noise distribution to the target distribution by directly regressing the velocity field, which enables generation through ODE-based sampling.

Specifically, we sample $\bm{z}_0\sim\mathcal{N}(\bm{0},\bm{I})$ and $t\sim\mathcal{U}[0,1]$, and construct a linear transport path:
\begin{equation}
    \bm{z}_t=(1-t)\bm{z}_{0}+t\bm{z}_1 .
\end{equation}
The conditional velocity network $\bm{v}_\theta$ takes $(\bm{z}_t,t,\bm{c})$ as input and predicts the velocity field that moves $\bm{z}_t$ toward the target detail latent. The standard FM objective is:
\begin{equation}
    \mathcal{L}_{\mathrm{FM}}
    =
    \left\|
    \bm{v}_\theta(\bm{z}_t,t,\bm{c})
    -(\bm{z}_1-\bm{z}_0)
    \right\|_2^2 .
\end{equation}

To improve the stability of large-step inference, we further introduce a shortcut consistency constraint~\cite{shortcut}. Given a step size $\Delta t$, we estimate an intermediate state by one short-step update:
\begin{equation}
    \bm{z}_{t+\Delta t}^{\prime}
    =
    \bm{z}_t+
    \bm{v}_{\theta}(\bm{z}_t,t,\bm{c},\Delta t)\Delta t .
\end{equation}
The average velocity over two consecutive short steps is computed as:
\begin{equation}
    \label{eq:shortcut_target}
    \bar{\bm{v}}_\theta
    =
    \frac{1}{2}
    \left[
        \bm{v}_\theta(\bm{z}_t,t,\bm{c},\Delta t)
        +
        \bm{v}_\theta(\bm{z}_{t+\Delta t}^{\prime},t+\Delta t,\bm{c},\Delta t)
        \right] .
\end{equation}
We then encourage the velocity predicted for a single large step $2\Delta t$ to be consistent with the average velocity of the two short steps:
\begin{equation}
    \mathcal{L}_{\mathrm{SC}}
    =
    \left\|
    \bm{v}_\theta(\bm{z}_t,t,\bm{c},2\Delta t)
    -
    \bar{\bm{v}}_\theta
    \right\|_2^2 .
\end{equation}

In each training batch, one quarter of the samples are optimized with the shortcut consistency objective, while the remaining samples use the standard FM objective.

\subsubsection{One-Step ODE Solver}

During inference, we sample an initial state $\bm{z}_0 \sim \mathcal{N}(\bm{0}, \bm{I})$ and generate the detail latent by solving the flow ODE from $t=0$ to $t=1$:
\begin{equation}
    \frac{\mathrm{d}\bm{z}}{\mathrm{d}t}
    =
    \bm{v}_\theta(\bm{z}, t, \bm{c}),
    \quad
    \bm{z}(0)=\bm{z}_0 .
\end{equation}
Benefiting from the shortcut consistency training, we approximate this transport with a single Euler step:
\begin{equation}
    \hat{\bm{z}}_1
    =
    \bm{z}_0
    +
    \bm{v}_\theta(\bm{z}_0,0,\bm{c},1).
\end{equation}
The resulting $\hat{\bm{z}}_1$ is used as the generated detail latent for subsequent continuous-scale Gaussian rendering.

\subsection{Continuous-Scale Gaussian Rendering}
\label{sec:continuous-scale-rendering}
\subsubsection{Scale-Aware Feature Fusion}
Given the generated detail latent $\hat{\bm{z}}_1$, the LR input $I_{\mathrm{LR}}$, and the target scale factor $s$, we construct a scale-aware feature field:
\begin{equation}
    \bm{F} =\mathcal{D}(\hat{\bm{z}}_1,I_{\mathrm{LR}},s)\in\mathbb{R}^{s\cdot H\times s\cdot W\times C},
\end{equation}
where $\mathcal{D}(\cdot)$ is a feature fusion decoder that integrates the generated detail latent with LR structural features under scale modulation. The output feature field $\bm{F}$ is used for subsequent Gaussian parameter prediction and continuous rendering. We implement $\mathcal{D}(\cdot)$ with a multi-stage feature refinement design following~\cite{ldcsr}.

\subsubsection{Continuous Rendering}

Given the fused feature field $\bm{F}$, we assign a Gaussian kernel to each spatial location from a predefined Gaussian kernel bank. Specifically, the kernel bank contains $K$ learnable Gaussian candidates:
\begin{equation}
    \mathcal{B}=\{(\bm{\tilde{\Sigma}}^{(k)},o^{(k)})\}_{k=1}^{K},
\end{equation}
where $\bm{\tilde{\Sigma}}^{(k)}$ and $o^{(k)}$ denote the covariance matrix and opacity of the $k$-th kernel, respectively. For each spatial location $i$, its coordinate determines the Gaussian center $\bm{\mu}_i$, while the corresponding fused feature $\bm{F}_i$ is used to predict the assignment weight over the kernel bank:
\begin{equation}
    \bm{\pi}_i=\mathrm{Softmax}(P(\bm{F}_i)) \in \mathbb{R}^{K},
\end{equation}
where $P(\cdot)$ is a lightweight prediction head.

The final Gaussian kernel at location $i$ is then obtained by weighting the predefined candidates:
\begin{equation}
    \bm{\Sigma}_i=\sum_{k=1}^{K}\pi_{i,k}\bm{\tilde{\Sigma}}^{(k)},
    \quad
    o_i=\sum_{k=1}^{K}\pi_{i,k}o^{(k)} .
\end{equation}
The Gaussian center $\bm{\mu}_i$ is determined by the spatial coordinate of location $i$, and the feature vector $\bm{v}_i$ is taken from the fused feature field $\bm{F}$. Thus, each location is represented as a Gaussian primitive $(\bm{\mu}_i,\bm{\Sigma}_i,o_i,\bm{v}_i)$.

For any query location $x_q$, the spatial response of the $i$-th Gaussian kernel is computed as:
\begin{equation}
    f_i(x_q \mid \bm{\mu}_i, \bm{\Sigma}_i) = \frac{1}{2\pi |\bm{\Sigma}_i|^{1/2}} e^{( -\frac{1}{2} (x_q - \bm{\mu}_i)^\top \bm{\Sigma}_i^{-1} (x_q - \bm{\mu}_i))}.
\end{equation}

The continuous feature at $x_q$ is rendered by the 2D-Gaussian aggregation:
\begin{equation}
    \mathcal{C}_{\mathrm{2DGS}}(x_q \mid \bm{\mu}, \bm{\Sigma}, o, \bm{v})
    =
    \sum_i
    f_i(x_q \mid \bm{\mu}_i, \bm{\Sigma}_i)\cdot \bm{c}_i,
\end{equation}
where $\bm{c}_i=o_i \cdot \bm{v}_i$ denotes the opacity-modulated feature contribution of the $i$-th Gaussian.

Based on the aggregated feature, a lightweight residual predictor $M(\cdot)$ estimates the residual:
\begin{equation}
    \Delta I(x_q)
    =
    M\left(
    \mathcal{C}_{\mathrm{2DGS}}(x_q \mid \bm{\mu}, \bm{\Sigma}, o, \bm{v})
    \right).
\end{equation}
The final super-resolved output is obtained by adding the predicted residual to the bicubic-upsampled LR image:
\begin{equation}
    I_{\mathrm{SR}}(x_q)=I_{\mathrm{LR}}^{\uparrow}(x_q)+\Delta I(x_q).
\end{equation}

During training, the encoder in \eqref{eq:encoder} produces the ground-truth detail latent $\bm{z}_1$ from the HR-LR pair, and we jointly optimize the encoder and the Gaussian rendering module with
\begin{equation}
    \mathcal{L}
    =
    \| I_{\mathrm{SR}} - I_{\mathrm{HR}} \|_1
    +
    \lambda_{1} L_{\mathrm{LPIPS}}
    +
    \lambda_{2} L_{\mathrm{adv}}
    +
    \lambda_{3} L_{\mathrm{KL}},
\end{equation}
where the $\ell_1$ term enforces reconstruction fidelity, $L_{\mathrm{LPIPS}}$ improves perceptual consistency, $L_{\mathrm{adv}}$ improves perceptual realism~\cite{ldm, esser2021taming}, and $L_{\mathrm{KL}}$ regularizes the latent distribution; $\lambda_{1}$, $\lambda_{2}$, and $\lambda_{3}$ are the corresponding weights. During inference, the unavailable $\bm{z}_1$ is replaced by the FM-predicted detail latent $\hat{\bm{z}}_1$.

\section{Experiments and Analysis}
\subsection{Datasets}
We evaluated our method on three public RS datasets: AID~\cite{aid_dataset}, DOTA~\cite{dota_dataset}, and DIOR~\cite{dior_dataset}. AID comprises 10{,}000 HR images of size $600\times600$ spanning 30 aerial scene categories. We partition it into 8{,}000 training, 1{,}000 validation, and 1{,}000 test images. For cross-dataset evaluation, we randomly selected 500 images each from DOTA and DIOR as test sets, with dimensions of $512\times512$ and $800\times800$, respectively. For efficient ablation evaluation, we construct AID-tiny by randomly selecting 30 test images from AID and center-cropping them to $512\times512$.

\subsection{Implementation Details}
The continuous rendering module uses a Gaussian kernel bank with $K=100$ Gaussian candidate kernels and a $7\times7$ kernel splatting region to predict the residual $\Delta I$. The velocity network in the conditional flow matching module adopts a \mbox{U-Net} architecture following LDCSR~\cite{ldcsr}.

First, we pretrain the continuous-scale Gaussian rendering module for 1{,}000 epochs with a batch size of 4 and a learning rate of $4.5\times10^{-6}$. Each epoch comprises 1{,}000 iterations. We set the LPIPS, adversarial, and KL weights to $\lambda_{1}=1$, $\lambda_{2}=0.5$, and $\lambda_{3}=1\times10^{-6}$, respectively, and select the best epoch checkpoint for next-stage training.

Next, we train the FM-based detail latent generation module for a total of 400 epochs with a batch size of 8, a learning rate of $1\times10^{-5}$, and 1{,}000 warmup steps. The exponential moving average is maintained with a decay rate of 0.999, and its weights are used to compute the shortcut target in \eqref{eq:shortcut_target}. All reported results are obtained using single-step inference.

Both stages perform online bicubic downsampling from AID HR images, with the scale factor $s$ uniformly sampled from $[1,\,8]$ and shared across each mini-batch. Standard augmentations including random horizontal and vertical flipping and $90^\circ$ rotation are applied. All experiments are implemented in PyTorch on a single NVIDIA RTX 4090 GPU.

\begin{table*}[t]
  \centering
  \footnotesize
  \caption{Continuous-scale SR comparison on AID and DOTA. Each cell reports FID\,$\downarrow$ / LPIPS\,$\downarrow$ at the corresponding scale factor. \best{Best} and \second{Second-best} results are highlighted.}
  \label{tab:cssr-results}
  \renewcommand\arraystretch{1.1}
  \setlength{\tabcolsep}{5.5pt}
  \begin{tabular}{l|cccc|cccc}
  \Xhline{3\arrayrulewidth}
  \multirow{2}{*}{Method}
  & \multicolumn{4}{c|}{AID}
  & \multicolumn{4}{c}{DOTA}\\
  \cline{2-5}
  \cline{6-9}
  & $\times$2.6
  & $\times$3.4
  & $\times$6.0
  & $\times$10.0
  & $\times$2.6
  & $\times$3.4
  & $\times$6.0
  & $\times$10.0\\
  \hline

  LIIF \cite{liif}
  & \makecell{9.22 / 0.200}
  & \makecell{27.49 / 0.283}
  & \makecell{80.88 / 0.460}
  & \makecell{136.70 / 0.595}
  & \makecell{16.53 / 0.133}
  & \makecell{43.22 / 0.203}
  & \makecell{100.55 / 0.355}
  & \makecell{160.71 / 0.486}\\

  CiaoSR \cite{ciaosr}
  & \makecell{8.53 / 0.188}
  & \makecell{24.98 / 0.268}
  & \makecell{75.01 / 0.432}
  & \makecell{132.34 / 0.559}
  & \makecell{14.85 / 0.126}
  & \makecell{38.80 / 0.190}
  & \makecell{94.65 / 0.332}
  & \makecell{151.92 / 0.448}\\

  LDCSR \cite{ldcsr}
  & \makecell{\second{6.92} / \best{0.087}}
  & \makecell{\second{13.57} / \best{0.137}}
  & \makecell{\second{33.86} / \second{0.257}}
  & \makecell{\second{82.31} / \second{0.465}}
  & \makecell{\best{11.55} / \best{0.069}}
  & \makecell{\second{25.68} / \best{0.114}}
  & \makecell{\second{61.93} / \second{0.249}}
  & \makecell{\second{129.71} / 0.460}\\

  GaussianSR \cite{gaussiansr}
  & \makecell{8.99 / 0.193}
  & \makecell{27.00 / 0.277}
  & \makecell{79.59 / 0.457}
  & \makecell{129.06 / 0.597}
  & \makecell{15.78 / 0.128}
  & \makecell{42.03 / 0.199}
  & \makecell{99.85 / 0.358}
  & \makecell{155.32 / 0.497}\\

  PMRF \cite{pmrf}
  & \makecell{20.99 / 0.160}
  & \makecell{25.10 / 0.193}
  & \makecell{36.48 / 0.268}
  & \makecell{84.04 / 0.484}
  & \makecell{50.91 / 0.162}
  & \makecell{55.34 / 0.191}
  & \makecell{73.29 / 0.259}
  & \makecell{130.40 / \second{0.436}}\\

  FlowGS (Ours)
  & \makecell{\best{6.59} / \second{0.106}}
  & \makecell{\best{13.56} / \second{0.159}}
  & \makecell{\best{31.94} / \best{0.256}}
  & \makecell{\best{61.57} / \best{0.452}}
  & \makecell{\second{11.88} / \second{0.078}}
  & \makecell{\best{25.46} / \second{0.126}}
  & \makecell{\best{59.16} / \best{0.222}}
  & \makecell{\best{106.61} / \best{0.398}}\\

  \Xhline{3\arrayrulewidth}
  \end{tabular}
\end{table*}

\begin{table}[!t]
  \centering
  \footnotesize
  \caption{Fixed-scale SR comparison on DIOR. Columns report LPIPS\,$\downarrow$ and FID\,$\downarrow$ per scale. \best{Best} and \second{Second-best} results are marked separately within fixed- and continuous-scale methods.}
  \setlength{\tabcolsep}{5.5pt}
  \renewcommand\arraystretch{1.1}
  \begin{tabular}{l|cc|cc|cc}
  \Xhline{3\arrayrulewidth}
  \multirow{2}{*}{Method}
  & \multicolumn{2}{c|}{$\times$2}
  & \multicolumn{2}{c|}{$\times$4}
  & \multicolumn{2}{c}{$\times$8}\\
  \cline{2-7}
  & LPIPS & FID & LPIPS & FID & LPIPS & FID\\
  \hline

  \multicolumn{7}{l}{\textit{Fixed-scale methods}}\\
  \hline
  HAT-L \cite{hat}
  & 0.118 & \second{7.40}
  & 0.337 & 32.68
  & 0.548 & 120.20\\

  SR3 \cite{sr3}
  & 0.069 & 8.24
  & \second{0.268} & \best{25.55}
  & \second{0.433} & \best{79.40}\\

  EDiffSR \cite{ediffsr}
  & \second{0.056} & 9.96
  & 0.332 & 32.46
  & 0.459 & \second{79.95}\\

  SPSR \cite{spsr}
  & \best{0.047} & 9.39
  & \best{0.203} & \second{30.60}
  & \best{0.405} & 82.82\\

  TTST \cite{ttst}
  & 0.113 & \best{7.22}
  & 0.332 & 31.59
  & 0.528 & 120.08\\
  \hline

  \multicolumn{7}{l}{\textit{Continuous-scale methods}}\\
  \hline
  LIIF \cite{liif}
  & 0.123 & \second{7.86}
  & 0.349 & 34.38
  & 0.561 & 124.24\\

  CiaoSR \cite{ciaosr}
  & 0.118 & \best{7.36}
  & 0.334 & 32.17
  & 0.535 & 117.08\\

  LDCSR \cite{ldcsr}
  & \best{0.055} & 10.02
  & \second{0.212} & \second{23.03}
  & 0.387 & \best{67.31}\\

  GaussianSR \cite{gaussiansr}
  & 0.119 & 7.91
  & 0.345 & 34.28
  & 0.560 & 120.87\\

  PMRF \cite{pmrf}
  & 0.161 & 34.29
  & 0.243 & 41.63
  & \second{0.361} & 72.97\\

  FlowGS (Ours)
  & \second{0.068} & 8.70
  & \best{0.205} & \best{21.87}
  & \best{0.348} & \second{71.36}\\
  \Xhline{3\arrayrulewidth}
  \end{tabular}
  \label{tab:fssr-results}
\end{table}

\subsection{Comparison With State-of-the-Art Methods}

We compare FlowGS with both continuous-scale and fixed-scale SR methods. For continuous-scale SR, we include three regression-based approaches, i.e., LIIF~\cite{liif}, CiaoSR~\cite{ciaosr}, and GaussianSR~\cite{gaussiansr}, as well as two generative methods, i.e., LDCSR~\cite{ldcsr} and PMRF~\cite{pmrf}. For fixed-scale SR, we compare against transformer-based models, including HAT-L~\cite{hat} and TTST~\cite{ttst}, diffusion-based methods, including SR3~\cite{sr3} and EDiffSR~\cite{ediffsr}, and the GAN-based method SPSR~\cite{spsr}. Each fixed-scale model is independently trained for each scale factor. We adopt LPIPS and FID as the main evaluation metrics. LPIPS measures perceptual similarity using deep feature representations and correlates better with human visual perception, while FID evaluates the distributional discrepancy between generated results and real HR images in the Inception feature space.

As shown in Table~\ref{tab:cssr-results}, FlowGS achieves the best FID in most settings on both AID and DOTA, indicating superior distributional realism. The gains are particularly clear at large upscaling factors, where the reconstruction problem becomes more ill-posed. Beyond continuous-scale SR, we further compare FlowGS with representative fixed-scale SR methods in Table~\ref{tab:fssr-results}. FlowGS achieves comparable or even superior performance to these fixed-scale counterparts at $\times4$ and $\times8$.

\begin{figure}[t]
    \centering
    \includegraphics[width=\linewidth]{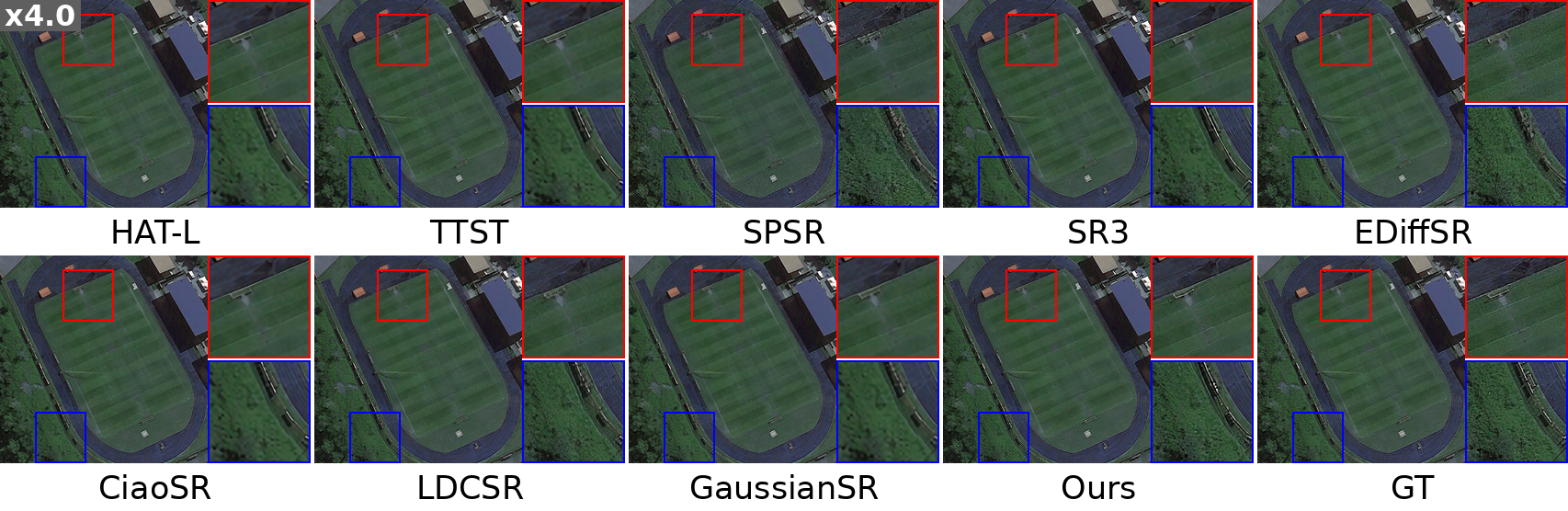}

    \vspace{0.6em}

    \includegraphics[width=\linewidth]{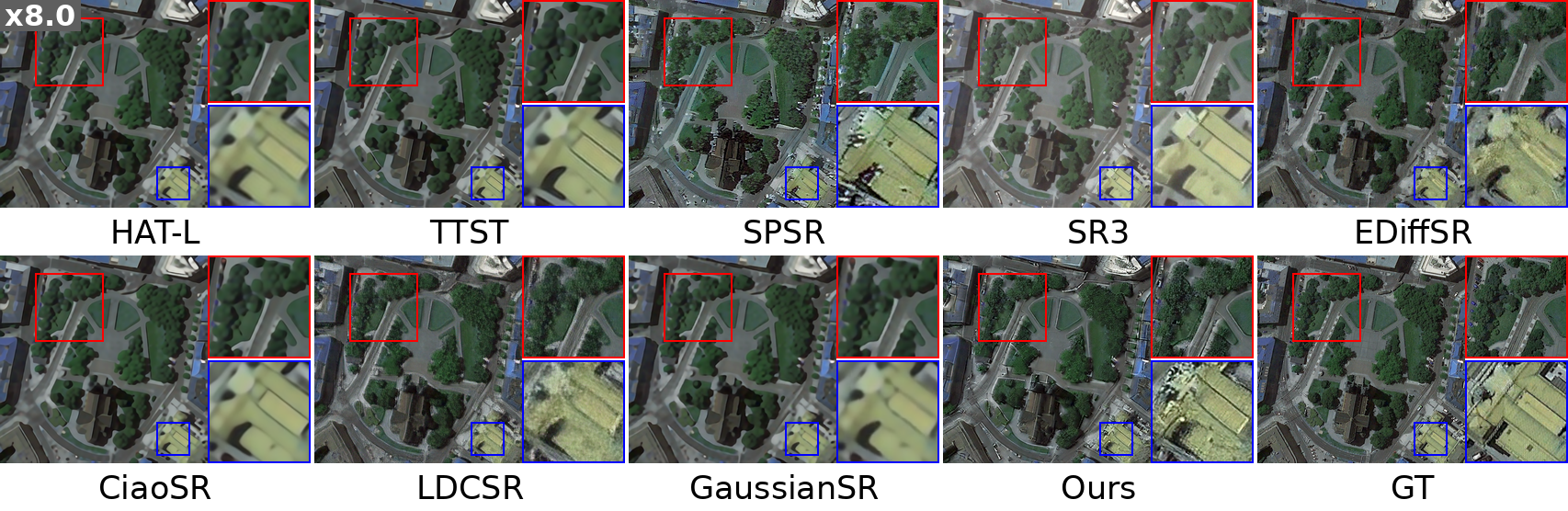}
    \caption{Qualitative comparison of continuous-scale SR methods on the AID dataset at scale factors of $\times4$ and $\times8$.}
    \label{fig:main-continuous-scale}
\end{figure}

From the visual results in Fig.\,\ref{fig:main-continuous-scale}, the differences among methods are modest at $\times$4.0 but become more evident at $\times$8.0. Regression-based methods generally produce smoother results with limited texture recovery, while generative methods such as SPSR, EDiffSR, and LDCSR restore more details but may still exhibit mild artifacts or local blur. In contrast, FlowGS produces sharper edges and more natural textures, achieving visual fidelity closer to the ground truth.

To further evaluate computational efficiency, we compare the inference time and FID of different methods in Fig.~\ref{fig:efficiency-FID}. FlowGS achieves a favorable efficiency--quality trade-off, requiring only a small fraction of the inference time of diffusion-based methods while maintaining better FID. This advantage is mainly attributed to shortcut consistency and the efficient Gaussian rendering process for continuous reconstruction.

\subsection{Ablation Studies}

We conduct ablation studies on two key designs of FlowGS, namely the FM-based detail latent generation and shortcut consistency, as reported in Table~\ref{tab:abl_main}. For a fair comparison, each variant is retrained under the same experimental setting. First, removing the FM-based detail latent generation module, denoted as \textit{w/o Flow}, causes a noticeable degradation in FID, suggesting that this module mainly enhances distributional realism rather than pixel-level fidelity. Second, replacing the bootstrap self-consistency target with a pure FM loss (\textit{w/o SC}) leads to a clear performance drop, indicating that shortcut consistency is crucial for enabling single-step inference.

\begin{table}[t]
  \centering
  \footnotesize
  \caption{Component ablation on AID-tiny under single-step inference.}
  \label{tab:abl_main}
  \setlength{\tabcolsep}{7pt}
  \renewcommand\arraystretch{1.1}
  \begin{tabular}{l|cc|cc|cc}
  \Xhline{3\arrayrulewidth}
  \multirow{2}{*}{Variant}
  & \multicolumn{2}{c|}{Component}
  & \multicolumn{2}{c|}{$\times$4}
  & \multicolumn{2}{c}{$\times$8}\\
  \cline{2-7}
  & Flow & SC
  & LPIPS & FID
  & LPIPS & FID\\
  \hline
  FlowGS
  & \checkmark & \checkmark
  & 0.195 & \best{19.43}
  & \best{0.333} & \best{53.23}\\

  w/o Flow
  & $\times$ & --
  & \best{0.174} & 22.30
  & 0.359 & 58.95\\

  w/o SC
  & \checkmark & $\times$
  & 0.185 & 23.29
  & 0.359 & 58.38\\

  \Xhline{3\arrayrulewidth}
  \end{tabular}
\end{table}

\begin{figure}[t]
    \centering
    \includegraphics[width=\linewidth]{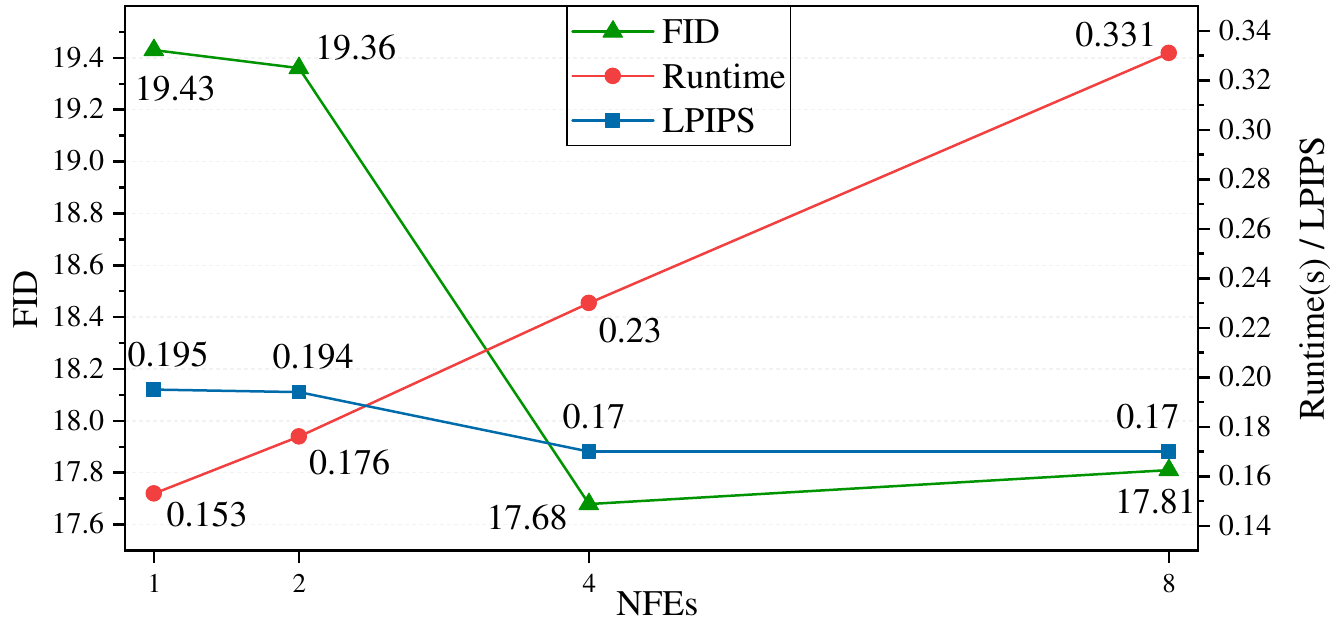}
    \caption{Quality--efficiency trade-off on AID at $\times4$ under different NFEs.}
    \label{fig:abl_steps}
\end{figure}

We further analyze the impact of inference steps, measured by the number of function evaluations (NFE), in Fig.~\ref{fig:abl_steps}. As NFE increases, perceptual quality improves in a stepwise manner and tends to saturate after 8 steps. Given that single-step inference achieves a favorable balance between quality and efficiency, we adopt it as the default setting.

\section{Conclusion}

We propose FlowGS, a generative reconstruction framework for arbitrary-scale SR in RS images. FlowGS enables single-step inference by generating high-frequency detail priors in latent space via FM with shortcut consistency. It further constructs continuous feature fields using 2D Gaussian splatting, allowing continuous reconstruction at arbitrary scales. Extensive experiments on multiple satellite datasets demonstrate that FlowGS achieves superior perceptual quality with high inference efficiency. In future work, we will further explore continuous-scale modeling and real-world degradation modeling to improve the robustness and generalization of FlowGS in practical RS applications.

\bibliographystyle{IEEEtran}
\bibliography{refs}

\end{document}